\documentclass{article}

\usepackage{PRIMEarxiv}

\usepackage[utf8]{inputenc} 
\usepackage[T1]{fontenc}    
\usepackage{hyperref}       
\usepackage{url}            
\usepackage{booktabs}       
\usepackage{amsfonts}       
\usepackage{nicefrac}       
\usepackage{microtype}      
\usepackage{lipsum}
\usepackage{fancyhdr}       

\usepackage{natbib}

\author{
  Landi He \\
  Shenzhen University of Advanced Technology \\
  \And
  Mingde Yao \\
  CUHK MMLab, CPII under InnoHK \\
  \And
  Shawn Young \\
  Shenzhen University of Advanced Technology \\
  \And
  Lijian Xu  \thanks{Corresponding author} \\
  Shenzhen University of Advanced Technology \\
  \texttt{xulijian@suat-sz.edu.cn}
}

\usepackage{times}
\usepackage{latexsym}

\usepackage{inconsolata}

\usepackage{graphicx}
\usepackage{subcaption}

\usepackage{amsmath}
\usepackage{amssymb}
\usepackage{amsthm}

\usepackage{xcolor}
\usepackage{colortbl}
\definecolor{rowblue}{rgb}{0.90,0.93,0.98}
\definecolor{rowgray}{rgb}{0.93,0.93,0.93}
\theoremstyle{remark}

\usepackage[capitalize]{cleveref}

\usepackage{makecell}
\usepackage{arydshln}

\title{Beyond Surrogate Gradients: Fully Differentiable Token Pruning for Vision-Language Models}

\begin{document}
\maketitle

\begin{abstract}
Visual token pruning reduces the computational cost of Vision-Language Models (VLMs) by removing redundant visual tokens. Existing methods typically rely on Gumbel-Softmax to approximate discrete selection during training. However, the optimization is driven by surrogate gradients rather than the true selection process, leading to unreliable learning of token importance. In this paper, we propose \textbf{DiffPrune}, which reformulates pruning as \textbf{continuous control of token information} instead of discrete selection learning. Specifically, we introduce an Information Throttler that modulates each token using variance-preserving noise conditioned on importance scores, where higher scores induce less information suppression during training. This design directly operates on token representations, naturally providing a \textbf{fully differentiable} optimization path for learning token importance. At inference, tokens are removed via hard thresholding on the learned scores. Across ten VLM benchmarks, DiffPrune retains $96.5\%$ of full-model accuracy while accelerating LLM prefill by $2.85\times$, with only $0.69$ ms of inference overhead.
\end{abstract}

\section{Introduction}
\label{sec:intro}

Modern vision-language models (VLMs) suffer substantial inference costs due to the large number of visual tokens introduced by high-resolution images and videos \cite{arefeen2024vita}. To improve efficiency, recent studies have explored various approaches, including token compression, adaptive computation, and visual-token pruning \cite{vasu2025fastvlm,dong2025mmtok,feng2026see}. Among them, visual-token pruning offers a structurally direct form of efficiency improvement, as it reduces sequence length without introducing additional latent representations or auxiliary decoding stages\cite{jeddi2025similarity}. However, effective pruning requires a reliable token importance scorer, which is non-trivial in practice.

\begin{figure}[h]
    \centering
    \includegraphics[width=\linewidth]{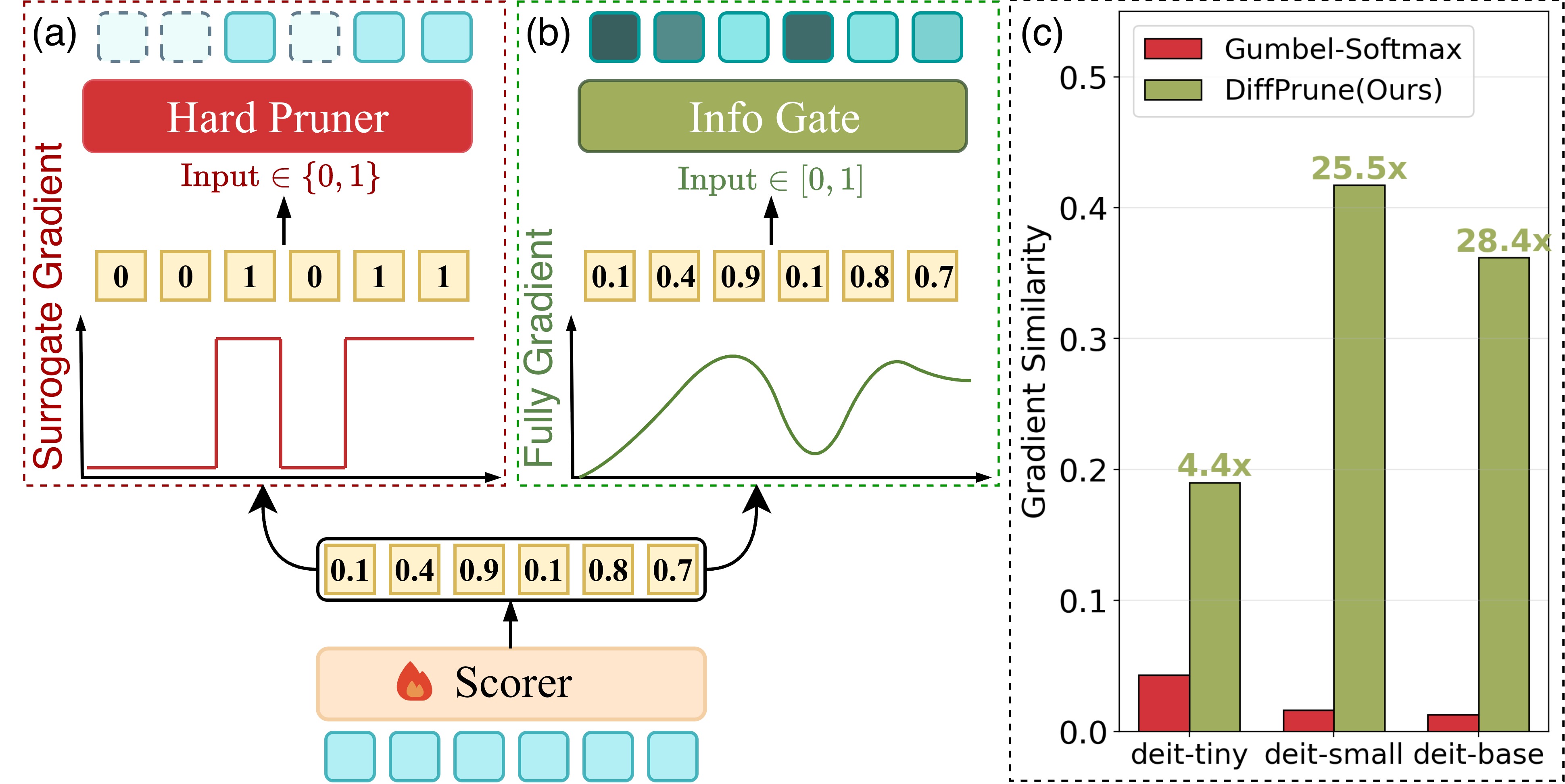}
    \caption{(a)\&(b) Gumbel-Softmax-style pruning vs. DiffPrune. (c) DiffPrune demonstrates more coherent direction on DeiT backbone~\cite{touvron2021training}, which enables fully differentiable optimization.}
    \label{fig:overview}
\end{figure}

Existing visual-token pruning methods can be broadly divided into training-free and training-based approaches depending on whether the token importance scorer is learned. Training-free methods use heuristic signals from pretrained backbones (e.g., attention or feature redundancy) to estimate token importance \cite{fastv,sparsevlm,tome,visionzip}, requiring no additional training and enabling direct inference-time deployment, but they are often misaligned with downstream tasks and lack adaptability across models \cite{divprune, zhang2026beyond}.
In contrast, training-based methods \cite{wang2023efficientvlm,wang2025bi,takezoe2026learnpruner} learn a scorer to estimate token importance. To enable optimization, they rely on relaxations such as Gumbel-Softmax (GS) or straight-through estimators (STE), which replace the discrete selection process with a differentiable surrogate. However, this surrogate does not match the true discrete pruning operation, so the training signal is based on surrogate gradients rather than the real pruning objective.

As shown in \Cref{fig:overview}(a), existing pipelines convert continuous importance scores into binary keep-or-drop decisions under a fixed budget, making the pruning operation inherently non-differentiable. GS and STE are used to approximate gradients through this discrete process, but they optimize a relaxed continuous surrogate rather than the original discrete selection problem \cite{zhao2025fine,liang2025dynamic}.
This mismatch between the surrogate objective and the true pruning objective leads to unstable optimization dynamics \cite{yang2026improving}. As illustrated in \Cref{fig:overview}(c), the resulting gradients exhibit high variance across training iterations, since small changes in importance scores can induce abrupt changes in the selected token subset. This effect becomes more pronounced as model size increases, where more tokens lie near the selection boundary.

In contrast, our method, as illustrated in  \Cref{fig:overview} (b), avoids optimizing through discrete relaxation. Instead, it maintains a continuous importance formulation throughout training and directly modulates token information flow under a fixed budget constraint. This yields a smoother optimization landscape and significantly improves gradient consistency. Importantly, our formulation remains fully differentiable throughout training, leading to substantially higher gradient alignment (28.4× on DeiT-Base).

In this paper, we propose DiffPrune, a fully differentiable  visual-token pruning framework for VLMs.
During training, every operator on the path from scorer logits to language-model loss has a backward gradient given by the analytic derivative of the same continuous function used in the forward pass.
The gradient path uses no surrogate gradients, argmax, hard selection, or straight-through estimators.
During training, DiffPrune does not make binary keep-or-drop decisions, but instead changes the amount of information carried by each visual token in a continuous way.
A Soft Top-$K$ head polarises the scorer logits into continuous weights and constrains their total budget to match hard top-$K$.
The Information Throttler has two components.
First, a variance-preserving (VP) noise injector uses these weights to modulate each token.
Second, a trainable denoiser maps the noised tokens back to the representation space expected by the downstream LLM.
At inference, Soft Top-$K$ is replaced by hard top-$K$, the denoiser is removed, and only the scorer is added to the inference path.
The base VLM remains frozen throughout.

Contributions can be concluded as follows.
\begin{itemize}
    \item We show that the limitation of surrogate-gradient pruning lies at the operator level rather than the scorer level, and propose DiffPrune as a fully differentiable framework for visual-token pruning.
    \item We introduce VP-Noise, a diffusion-inspired Throttler that interpolates each token with Gaussian noise according to its retention weight, suppressing token content without relying on a discrete keep-or-drop operation.
    \item In controlled DeiT probes, DiffPrune reaches up to $28.4\times$ higher cross-batch gradient-direction coherence than Gumbel-Softmax; across LLaVA-1.5-7B, LLaVA-NEXT-7B, and Qwen2.5-VL-7B evaluations, it preserves high task performance under aggressive pruning while adding only $0.69$\,ms of inference overhead.
\end{itemize}

\section{Related Work}
\label{sec:related-work}

\subsection{Training-Free Visual Token Compression}
\label{sec:related-free}

Training-free compression keeps the VLM backbone fixed and score tokens using signals already present in the model.
We organize this line by the source of the token score \cite{shao2025survey,yao2026towards}.
Attention-based methods use saliency from the vision tower or the language decoder \cite{fastv,sparsevlm,fastervlm,vtw,hired,pyramiddrop,topv}.
Redundancy-based methods remove or merge visually similar features \cite{tome,prumerge,visionzip,fitprune,feather-throttle}.
Coverage-based methods keep tokens that are diverse or spatially well spread \cite{divprune,dart,holov,balanced-token-pruning,fangprune}.
These methods are strong plug-in baselines because they do not train a new selector.
While strong as plug-in baselines, these methods do not adapt their compression rules to downstream language-model loss. Learned pruning instread must convert continuous scores into discrete keep-or-drop decisions while enabling gradient flow.

\subsection{Learning-based- Token Selection}
\label{sec:related-based}

Learned pruning must train the steps.
A common solution is surrogate-gradient pruning.
The forward pass uses a discrete or nearly discrete mask, while the backward pass differentiates a smooth relaxation.
This pattern appears in semi-structured weight sparsity, KV-cache eviction, and text-side token pruning \cite{fang2024maskllm,zhou2026lkv,li2023constraint, wu2025vlm}.
These methods use different relaxations, including Gumbel-Softmax with STE, differentiable sorting, and hard-concrete gates, and compress different objects.
The core issue remains: the score-to-mask function in the forward pass is not the one used for backpropagation.

Learned visual-token pruning follows the same pattern.
DynamicViT \cite{rao2021dynamicvit} samples token survival with Gumbel-Softmax and passes gradients through the soft relaxation.
ATP-LLaVA \cite{atp-llava} uses a temperature-sharpened sigmoid-threshold mask.
Dynamic-LLaVA \cite{huang2025dynamic} and LightVLA \cite{jiang2025better} use binary or argmax-like selections with Gumbel-Softmax-style backward paths.
These methods differ in scorer architecture, pruning location, and budget control, but are similar in how continuous scores become selected tokens.

Recent variants point to the same bottleneck.
Shiva-DiT \cite{zhang2026shiva} replaces the softmax surrogate with a residual-aware STE in diffusion transformers.
TwigVLM++ \cite{twigvlmpp} trains a pruning head through distillation and policy-gradient reinforcement learning.
These routes either keep a discrete forward selection with a surrogate backward path, or avoid analytic differentiation through selection.
DiffPrune instead uses a continuous forward operator, whose backward gradients are the analytic derivatives of the same function.
\Cref{sec:pilot} formalises this contrast at the operator and loss-landscape levels.

\section{Pilot Studies}
\label{sec:pilot}

\subsection{Limitations of Gumbel-Softmax Pruning}
\label{sec:pilot-limits}

The main problem of surrogate-gradient pruning lies in the discrete selection operator itself, rather than only in the scorer design.
Let the logits output by the scorer be $s \in \mathbb{R}^{N}$ and Gumbel perturbation be $g$.
In the forward pass, the model generates a hard mask via top-$K$; in the backward pass, the gradient is usually passed directly from the mask $m$ back to the logits $s$:
\begin{align}
\mathrm{forward:}\quad & m = \mathrm{topK}(s + g) \in \{0,1\}^{N}, \nonumber\\
\mathrm{backward:}\quad & \frac{\partial L}{\partial s} \leftarrow \frac{\partial L}{\partial m}.
\label{eq:gs-ste}
\end{align}
The forward computation and the backward gradient here do not correspond to the same differentiable function.
The top-$K$ selection actually executed in the forward pass is piecewise constant with respect to $s$: as long as the ordering of the elements does not change, the retained token set remains unchanged, and therefore its derivative is almost everywhere zero \cite{shekhovtsov2020path,shah2024improving}.
Only when a score crosses the $K$-th order statistic and triggers an index swap does the output mask change discontinuously.
In contrast, the backward pass does not use the true Jacobian of this hard selection operator.
Instead, it approximates that Jacobian with the Jacobian of the identity map, directly passing $\partial L/\partial m$ to $s$.
Therefore, the gradient is not from the actual forward computation but a surrogate estimate, introducing a systematic gradient bias \cite{shekhovtsov2021bias} rather than unbiased random noise.

\subsection{A Gradient-Continuous Alternative}
\label{sec:pilot-alt}

Based on the above analysis, we propose an improved scheme: route every token through one analytic function whose value is the forward and whose Jacobian is the backward.
Rather than ask whether token $i$ is kept, we ask how much of its information is admitted, encode the answer as $\alpha_i \in [0,1]$, and apply a gradient-continuous Throttler $T(x_i;\alpha_i)$ to every token, with $\alpha$ supplied by a Soft Top-$K$ head under $\sum_i \alpha_i \approx K$; unlike the soft companion in \cref{eq:gs-ste}, Soft Top-$K$ is the operator the forward executes.
The probe below instantiates $T$ as a per-token scale composed with variance-preserving Gaussian mixing, with the design unpacked in \cref{sec:method}.

We visualize the scorer loss landscape with the filter-normalized
two-dimensional slice of Li et al.~\cite{li2018visualizing}. The probe inserts a
two-layer MLP scorer with roughly $14$M parameters between the patch embedding
layer and the frozen DeiT-base encoder. The patch embedding layer
outputs $N=196$ patch tokens, and the pruning ratio removes $70\%$ of them,
i.e., $K=58$. We compare Gumbel-softmax, which sends the hard top-$K$ tokens
to the encoder with an STE backward pass, against DiffPrune, which applies
Soft Top-$K$ scores through a continuous token throttler.

\begin{figure}[h]
  \centering
  \includegraphics[width=\columnwidth]{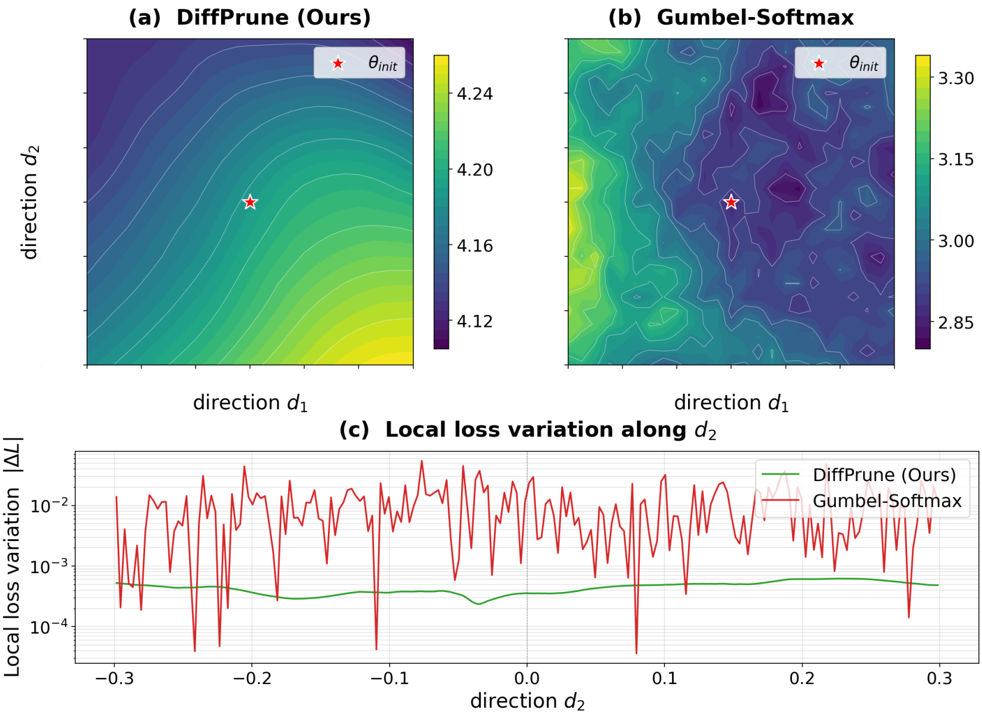}
  \caption{Gradient direction visualization. Our fully differentiable DiffPrune yields smoother and more coherent gradient directions than Gumbel-Softmax-based methods, facilitating more effective optimization.
}
  \label{fig:pilot}
\end{figure}

\Cref{fig:pilot}(a) shows that DiffPrune yields a smooth loss surface with
gradual contour changes. In contrast, \Cref{fig:pilot}(b) exhibits more
irregular loss contours with sharper local variations. \Cref{fig:pilot}(c)
makes this difference more explicit by plotting the stepwise loss variation
along the $d_2$ axis, measured as
$\lvert L(\beta_{i+1}) - L(\beta_i)\rvert$. The mean stepwise loss variation
of Gumbel-softmax is roughly $24\times$ that of DiffPrune.
These results suggest that the discontinuity introduced by hard token
selection and STE is not only a backward-pass mismatch, but also appears
directly in the scorer optimization landscape. By replacing discrete selection
with continuous throttling during training, DiffPrune provides a smoother
objective for optimizing the pruning scorer.

\section{Methodology}
\label{sec:method}

\subsection{Overview}
\label{sec:method-arch}

\begin{figure}[!htb]
  \centering
  \includegraphics[width=0.99\textwidth]{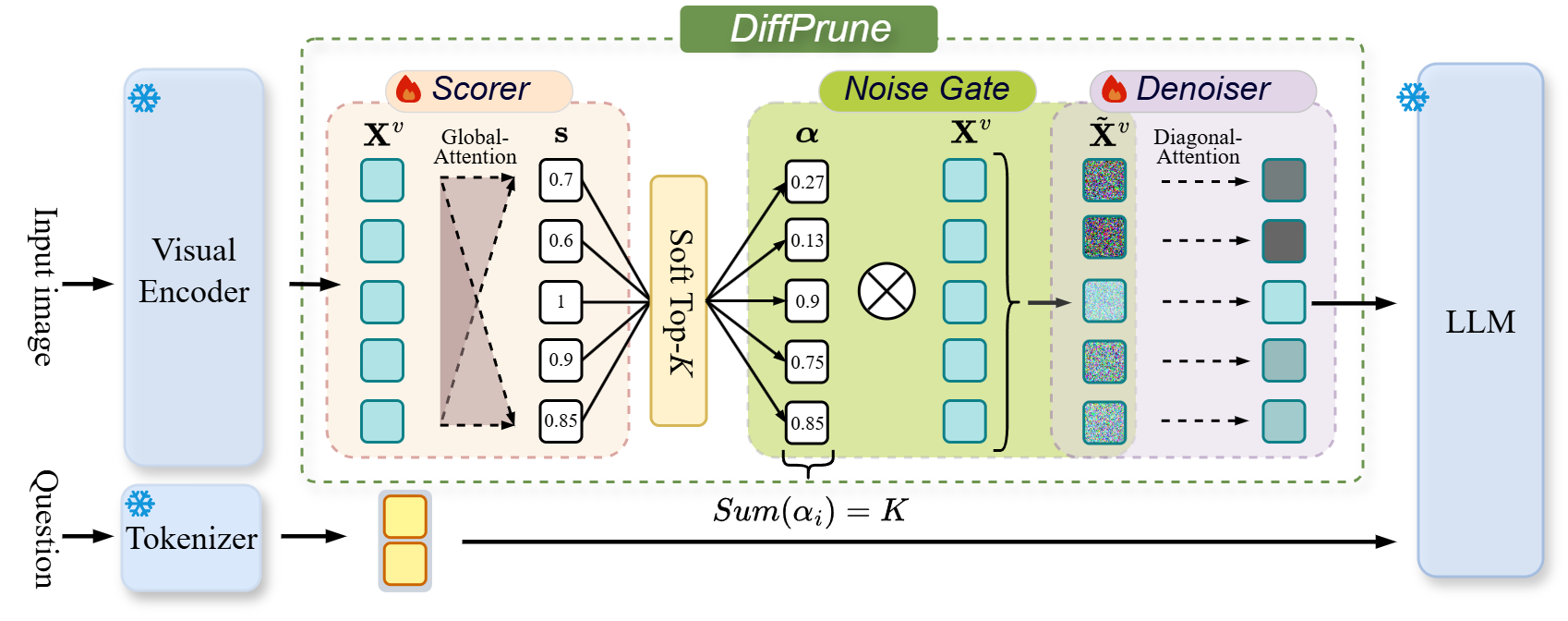}
  \caption{
Overview of \textbf{DiffPrune}. Visual tokens are first assigned importance scores by a \textbf{Scorer}. \textbf{Soft Top-$K$} transforms these scores into continuous retention weights satisfying a fixed budget constraint. Rather than making discrete keep-or-drop decisions, DiffPrune uses these weights to continuously throttle token information through a \textbf{Noise Gate}, where less important tokens receive stronger perturbations. A train-only \textbf{Denoiser} reconstructs the resulting representations before they are processed by the frozen LLM. At inference, DiffPrune collapses to standard hard top-$K$ token selection using the learned importance scores.
}
\label{fig:framework}
\end{figure}

DiffPrune builds a fully differentiable framework for visual-token pruning in VLMs.
\Cref{fig:framework} gives the overview: DiffPrune trains a token scorer through a continuous score-to-loss path and deploys it as a hard top-$K$ pruner.
Given an image $\mathbf{I}$, the frozen vision encoder produces visual tokens $\mathbf{X}^v=\mathcal{E}_v(\mathbf{I})\in\mathbb{R}^{N\times d_v}$.
The framework consists of a Scorer $\mathcal{S}_\theta$, a budgeted Soft Top-$K$ map $\Phi_K$, a token-wise Throttler $T$, and a train-only Denoiser $\mathcal{D}_\phi$.
The base VLM is frozen; only $\theta$ and $\phi$ are optimized.

Let $\mathbf{W}$ denote the text tokens, $p_\Theta$ the frozen language model, and $\mathbf{Z}$ the multimodal sequence consumed by it.
Training is written as
\begin{equation}
\begin{aligned}
\min_{\theta,\phi}\quad
& -\sum_t \log p_\Theta\!\left(y_t^* \mid y_{<t}^*, \mathbf{Z}\right),\\
\mathbf{Z}
&= [\,\mathbf{P}_{v\to\ell}(\bar{\mathbf{X}}^v);\,\mathcal{E}_t(\mathbf{W})\,],
\end{aligned}
\label{eq:train-fwd}
\end{equation}
where $\bar{\mathbf{X}}^v=\mathcal{D}_\phi(T(\mathbf{X}^v;\boldsymbol{\alpha}))$ and
$\boldsymbol{\alpha}=\Phi_K(\mathcal{S}_\theta(\mathbf{X}^v)/\tau)$ with $\sum_i\alpha_i=K$.
Here $K$ is the token budget, i.e., the target number of visual tokens retained by the deployed pruner; during training, it is represented as the total retention mass $\sum_i\alpha_i$.
Hard token selection is absent from \cref{eq:train-fwd}; the scorer is trained through the continuous operators that are actually executed in the forward pass.

At inference, DiffPrune uses the trained scorer as a ranking function and gathers the original top-$K$ visual tokens:
\begin{equation}
\hat{\mathbf{X}}^v
= \operatorname{TopK}\!\left(\mathbf{X}^v;\mathcal{S}_\theta(\mathbf{X}^v),K\right)
\in \mathbb{R}^{K\times d_v}.
\label{eq:infer-fwd}
\end{equation}
Here $\operatorname{TopK}(\mathbf{X};\mathbf{s},K)$ denotes score-indexed token gathering.
The retained tokens keep their original order and position indices.
The Throttler and Denoiser are removed from the deployed graph, leaving one scorer pass plus this gather operation.

\subsection{Scorer and Budgeted Soft Top-$K$}
\label{sec:method-softtopk}

The Scorer $\mathcal{S}_\theta$ is a two-layer Transformer encoder followed by a linear head that assigns one scalar logit $s_i$ to each visual token $\mathbf{x}_i^v$.
It predicts only an ordering signal; budget control is delegated to $\Phi_K$.
We then apply a budgeted Soft Top-$K$ map, a differentiable top-$K$ operator that returns continuous retention weights under a fixed cardinality budget \cite{struski2025lapsum}:
\begin{equation}
\boldsymbol{\alpha}
= \Phi_K\!\left(\mathbf{s}/\tau\right)\in[0,1]^N,
\qquad
\sum_{i=1}^{N}\alpha_i = K.
\label{eq:soft-topk}
\end{equation}
The budget is enforced by $\Phi_K$, so no auxiliary budget loss is used.
The temperature $\tau$ controls sharpness: cosine annealing lowers $\tau$ during training, making $\boldsymbol{\alpha}$ increasingly close to a binary top-$K$ mask.
With no score tie at the $K$-th boundary, the low-temperature limit matches the hard top-$K$ ranking used at inference in \cref{eq:infer-fwd}.
During training, the downstream VLM consumes $\boldsymbol{\alpha}$ itself, and the backward pass differentiates the same continuous map used in the forward pass.
This differs from the surrogate-gradient operator in \cref{eq:gs-ste}, where the forward executes a hard mask but the backward follows a different smooth path.

\subsection{Information Throttling with VP-Noise}
\label{sec:method-throttler}

Soft Top-$K$ produces continuous retention weights, but the downstream VLM must experience them as token-level information restriction.
DiffPrune therefore uses a token-wise Throttler $T(\mathbf{x}_i;\alpha_i)\mapsto\tilde{\mathbf{x}}_i$ for each visual token $\mathbf{x}_i\in\mathbb{R}^{d_v}$ and weight $\alpha_i\in[0,1]$.
A simple scale gate, $\tilde{\mathbf{x}}_i=\alpha_i\mathbf{x}_i$, is continuous but weak: low-scoring tokens keep their feature direction, and normalization layers can reduce the effect of magnitude shrinkage.

\begin{figure}[h]
  \centering
  \includegraphics[width=\columnwidth]{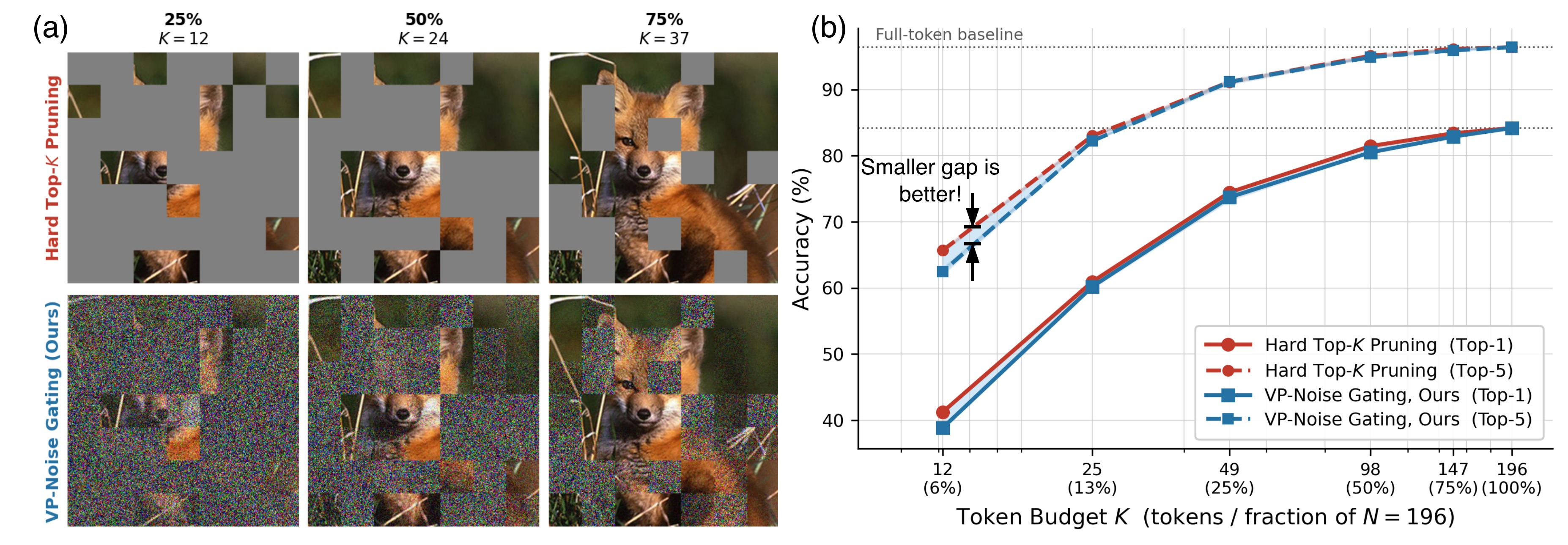}
  \caption{
    Our VP-Noise suppresses token information in a manner functionally equivalent to hard top-$K$ pruning, yet remains fully differentiable. (a) Visual comparison of VP-Noise and hard masking. (b) Their accuracy remains nearly identical under different pruning ratios, demonstrating the equivalence. }
  \label{fig:vp-validation}
\end{figure}

We therefore instantiate $T$ with a variance-preserving noising form inspired by the forward process of diffusion models \cite{ho2020denoising},
\begin{equation}
\tilde{\mathbf{x}}_i = \sqrt{\alpha_i}\,\mathbf{x}_i + \sqrt{1-\alpha_i}\,\boldsymbol{\epsilon}_i,
\qquad \boldsymbol{\epsilon}_i \sim \mathcal{N}(\mathbf{0},\mathbf{I}),
\label{eq:vp-noise}
\end{equation}
which we refer to as VP-Noise.
For high-scoring tokens, $\alpha_i\approx1$ preserves the original representation; for low-scoring tokens, $\alpha_i\approx0$ replaces most visual content with isotropic noise.
For normalized visual tokens, the square-root coefficients keep feature scale approximately stable while changing how much token content is available.
For fixed noise, this operation is differentiable in $\alpha_i$, so the Scorer is trained through the same noised representation consumed in the forward pass.
\Cref{fig:vp-validation} validates VP-Noise from complementary visual and quantitative views.
Panel (a) compares hard top-$K$ pruning and VP-Noise gating at $25\%$, $50\%$, and $75\%$ keep ratios, showing that VP-Noise keeps the retained spatial layout while corrupting low-retention patches instead of masking them out.
Panel (b) reports ImageNet-1K Top-1/Top-5 accuracy under random scores; the small gap to hard top-$K$ indicates that the continuous throttling path imposes comparable restriction strength.

\subsection{Diagonal-Attention Denoiser}
\label{sec:method-denoiser}

VP-Noise makes token removal differentiable, but low-retention tokens may deviate from the feature distribution expected by the frozen downstream stack.
We therefore insert a train-only Denoiser $\mathcal{D}_\phi$ after the Throttler to realign the throttled sequence while the scorer is being learned.

The Denoiser should not undo the information restriction imposed by VP-Noise.
DiffPrune implements it as a single Transformer block with an identity attention mask,
\[
\mathcal{D}_\phi(\tilde{\mathbf{X}}^v)
= \operatorname{Block}\!\left(\tilde{\mathbf{X}}^v;\operatorname{mask}=\mathbf{I}_N\right),
\]
so each token is transformed independently through projection, normalization, and feed-forward layers, without copying content from other positions.
The Denoiser is removed at inference.

\begin{table*}[!t]
  \centering
  \caption{
    \textbf{Results on LLaVA-1.5-7B.}
    We compare DiffPrune with recent visual-token pruning methods under three retained-token budgets.
    Loc.\ indicates where pruning is applied, and retention after normalizing each benchmark score by the unpruned upper bound. Dashed lines divide training-free (above) and training-based (below) methods within each block.
    \textcolor{red}{Best} and \textcolor{blue}{second-best} values are highlighted.
  }
  \label{tab:main-llava15}
  \scriptsize
  \setlength{\tabcolsep}{3pt}
  \resizebox{\linewidth}{!}{%
  \begin{tabular}{lcccccccccccc}
    \toprule
    Method & Loc. & GQA & MMB & MMB$_{\text{CN}}$ & MME & POPE & SQA & VQA$^{\text{v2}}$ & VQA$^{\text{Text}}$ & SEED & VizWiz & Avg. \\
    \midrule
    \rowcolor{rowgray} \multicolumn{13}{c}{\textit{Upper Bound, 576 Tokens (100\%)}} \\
    Vanilla & -- & 61.9 & 64.7 & 58.1 & 1862 & 85.9 & 69.5 & 78.5 & 58.2 & 60.5 & 54.3 & 100\% \\
    \midrule
    \rowcolor{rowgray} \multicolumn{13}{c}{\textit{Retain 192 Tokens (66.7\% pruning)}} \\
    FastV~{\scriptsize\textcolor{gray}{(ECCV'24)}} & L2--L32 & 52.7 & 61.2 & 53.5 & 1612 & 64.8 & 67.3 & 67.1 & 52.5 & 57.1 & 50.8 & 89.4\% \\
    HoloV~{\scriptsize\textcolor{gray}{(NeurIPS'25)}} & Pre & 59.0 & 65.4 & 58.0 & 1820 & 85.6 & 69.8 & 76.7 & 57.4 & -- & 50.9 & 98.2\% \\
    PRUNESID~{\scriptsize\textcolor{gray}{(ICLR'26)}} & Pre & 60.1 & 63.7 & -- & 1791 & 86.9 & 68.5 & 76.8 & 56.7 & 59.0 & 55.4 & \textcolor{red}{98.5}\% \\
    \hdashline[2pt/5pt]
    \hdashline[0pt/4pt]
    \hdashline[0pt/4pt]
    \hdashline[0pt/4pt]
    PyramidDrop~{\scriptsize\textcolor{gray}{(CVPR'25)}} & L8--L24 & 57.3 & 63.6 & 56.8 & 1797 & 82.3 & 69.2 & 75.1 & 56.5 & 54.7 & -- & 96.0\% \\
    VoCo-LLaMA~{\scriptsize\textcolor{gray}{(CVPR'25)}} & Pre & 61.4 & 56.3 & -- & 1596 & 84.5 & 66.6 & -- & 50.6 & 51.1 & -- & 91.1\% \\
    VisionZip~{\scriptsize\textcolor{gray}{(CVPR'25)}} & Pre & 60.1 & 63.4 & -- & 1834 & 84.9 & 68.2 & 77.4 & 57.8 & 57.1 & -- & 97.8\% \\
    
    \rowcolor{rowblue} \textbf{DiffPrune (Ours)} & Pre & 57.8 & 63.4 & 57.5 & 1791 & 86.5 & 70.2 & 76.6 & 55.3 & 59.2 & 55.9 & \textcolor{blue}{98.2}\% \\
    \midrule
    \rowcolor{rowgray} \multicolumn{13}{c}{\textit{Retain 128 Tokens (77.8\% pruning)}} \\
    FastV~{\scriptsize\textcolor{gray}{(ECCV'24)}} & L2--L32 & 49.6 & 56.1 & 55.9 & 1490 & 59.6 & 60.2 & 61.8 & 50.6 & 55.9 & 51.3 & 85.2\% \\
    HoloV~{\scriptsize\textcolor{gray}{(NeurIPS'25)}} & Pre & 57.7 & 63.9 & 56.5 & 1802 & 82.8 & 69.8 & 75.5 & 56.8 & -- & 51.5 & 96.8\% \\
    PRUNESID~{\scriptsize\textcolor{gray}{(ICLR'26)}} & Pre & 58.8 & 62.1 & -- & 1749 & 86.5 & 68.3 & 75.3 & 54.7 & 57.8 & 55.8 & \textcolor{blue}{96.9}\% \\
    \hdashline[2pt/5pt]
    \hdashline[0pt/4pt]
    \hdashline[0pt/4pt]
    \hdashline[0pt/4pt]
    PDrop~{\scriptsize\textcolor{gray}{(CVPR'25)}} & L8--L24 & 57.1 & 61.6 & 56.6 & 1761 & 82.3 & 68.4 & 72.9 & 56.6 & 53.3 & -- & 94.7\% \\
    VoCo-LLaMA~{\scriptsize\textcolor{gray}{(CVPR'25)}} & Pre & 61.5 & 56.4 & -- & 1640 & 84.5 & 66.6 & -- & 51.7 & 50.5 & -- & 91.5\% \\
    VisionZip~{\scriptsize\textcolor{gray}{(CVPR'25)}} & Pre & 58.9 & 62.6 & -- & 1823 & 83.7 & 68.3 & 76.6 & 56.8 & 55.8 & -- & 96.6\% \\

    \rowcolor{rowblue} \textbf{DiffPrune (Ours)} & Pre & 57.5 & 62.9 & 57.4 & 1765 & 85.7 & 70.2 & 76.1 & 54.9 & 58.4 & 56.1 & \textcolor{red}{97.6}\% \\
    \midrule
    \rowcolor{rowgray} \multicolumn{13}{c}{\textit{Retain 64 Tokens (88.9\% pruning)}} \\
    FastV~{\scriptsize\textcolor{gray}{(ECCV'24)}} & L2--L32 & 46.1 & 48.0 & 52.7 & 1256 & 48.0 & 51.1 & 55.0 & 47.8 & 51.9 & 50.8 & 76.8\% \\
    HoloV~{\scriptsize\textcolor{gray}{(NeurIPS'25)}} & Pre & 55.3 & 63.3 & 55.1 & 1715 & 80.3 & 69.5 & 72.8 & 55.4 & -- & 52.8 & 94.8\% \\
    PRUNESID~{\scriptsize\textcolor{gray}{(ICLR'26)}} & Pre & 57.1 & 58.8 & -- & 1733 & 83.8 & 67.8 & 73.7 & 54.2 & 56.1 & 56.9 & \textcolor{blue}{95.1}\% \\
    \hdashline[2pt/5pt]
    \hdashline[0pt/4pt]
    \hdashline[0pt/4pt]
    \hdashline[0pt/4pt]
    PDrop~{\scriptsize\textcolor{gray}{(CVPR'25)}} & L8--L24 & 47.5 & 58.8 & 50.5 & 1561 & 55.9 & 69.0 & 69.2 & 50.6 & 40.0 & -- & 82.7\% \\
    VoCo-LLaMA~{\scriptsize\textcolor{gray}{(CVPR'25)}} & Pre & 60.2 & 57.7 & -- & 1623 & 83.4 & 67.7 & -- & 51.1 & 50.0 & -- & 91.2\% \\
    VisionZip~{\scriptsize\textcolor{gray}{(CVPR'25)}} & Pre & 57.0 & 61.5 & -- & 1756 & 80.9 & 68.8 & 74.2 & 56.0 & 53.4 & -- & 94.2\% \\

    \rowcolor{rowblue} \textbf{DiffPrune (Ours)} & Pre & 56.8 & 62.6 & 56.6 & 1723 & 83.4 & 70.1 & 73.9 & 54.3 & 57.6 & 57.2 & \textcolor{red}{96.5}\% \\
    \bottomrule
  \end{tabular}%
  }
\end{table*}

\section{Experiments}
\label{sec:experiments}

\subsection{Experimental Setup}
\label{sec:exp-setup}

DiffPrune is trained on $10\%$ of the ImageNet-1K training split~\cite{deng2009imagenet} with captions from ImageNet-1K-VL-Enriched~\cite{visual_layer_imagenet1k_vl_enriched}.
Only the Scorer and Denoiser are optimized, adding approximately $84$M trainable parameters, while the base VLM remains frozen.
Complete implementation, training, baseline, and profiling details are provided in \cref{sec:appendix-impl}; benchmark definitions and reporting conventions are provided in \cref{sec:appendix-benchmarks}.

\subsection{Comparative Results}
\label{sec:main-results}

\paragraph{Results on LLaVA-1.5-7B.}
LLaVA-1.5-7B processes $336{\times}336$ inputs that produce $N{=}576$ visual tokens, and \cref{tab:main-llava15} reports three compression regimes that retain $192/128/64$ tokens (66.7\%/77.8\%/88.9\% pruning).
DiffPrune attains the strongest average retention at $K{=}128$ and $K{=}64$; against the closest prior method, PRUNESID, the gap moves from a $0.3\%$ deficit at $K{=}192$ to a $1.4\%$ lead at $K{=}64$, a direction consistent with the prediction of \cref{sec:pilot-limits} that tighter budgets amplify the gap between fully differentiable and surrogate-gradient routes.
The $K{=}192$ gap to PRUNESID should be read together with the efficiency analysis of \cref{sec:efficiency}, where the PRUNESID selection module is shown to cost roughly $60\times$ as much as ours.

\begin{table*}[!h]
  \centering
  \caption{
    \textbf{Results on LLaVA-NEXT-7B at $K{=}320$ (88.9\% pruning).}
    This high-resolution setting starts from $2880$ visual tokens; Avg.\ reports normalized average retention with \textcolor{red}{Best} and \textcolor{blue}{second-best} values highlighted.
  }
  \label{tab:main-llavanext}
  \scriptsize
  \setlength{\tabcolsep}{4pt}
  \resizebox{\linewidth}{!}{%
  \begin{tabular}{lcccccccccc}
    \toprule
    Method & Loc. & GQA & MMB & MMB$_{\text{CN}}$ & MME & POPE & SQA & VQA$^{\text{v2}}$ & VQA$^{\text{Text}}$ & Avg. \\
    \midrule
    \rowcolor{rowgray} \multicolumn{11}{c}{\textit{Upper Bound, 2880 Tokens (100\%)}} \\
    Vanilla & - & 64.2 & 67.4 & 60.6 & 1851 & 86.5 & 70.1 & 80.8 & 64.9 & 100\% \\
    \midrule
    \rowcolor{rowgray} \multicolumn{11}{c}{\textit{Retain 320 Tokens (88.9\% pruning)}} \\
    FastV~{\scriptsize\textcolor{gray}{(ECCV'24)}} & L2--L32 & 55.9 & 61.6 & 51.9 & 1661 & 71.7 & 62.8 & 71.9 & 55.7 & 87.6\% \\
    PDrop~{\scriptsize\textcolor{gray}{(CVPR'25)}} & L8--L24 & 56.4 & 63.4 & 56.2 & 1663 & 77.6 & 67.5 & 73.5 & 54.4 & 90.7\% \\
    DART~{\scriptsize\textcolor{gray}{(EMNLP'25)}} & L2 & 61.7 & 65.3 & 58.2 & 1710 & 84.1 & 68.4 & 79.1 & 58.7 & 95.6\% \\
    HiRED~{\scriptsize\textcolor{gray}{(AAAI'25)}} & Pre & 59.3 & 64.2 & 55.9 & 1690 & 83.3 & 66.7 & 75.7 & 58.8 & 93.4\% \\
    HoloV~{\scriptsize\textcolor{gray}{(NeurIPS'25)}} & Pre & 61.7 & 65.3 & 57.5 & 1738 & 83.9 & 68.9 & 79.5 & 58.7 & \textcolor{blue}{95.7}\% \\
    \rowcolor{rowblue} \textbf{DiffPrune (Ours)} & Pre & 62.3 & 64.7 & 57.4 & 1723 & 85.9 & 72.7 & 78.6 & 56.7 & \textcolor{red}{96.1}\% \\
    \bottomrule
  \end{tabular}%
  }
\end{table*}

\paragraph{Results on LLaVA-NEXT-7B.}
LLaVA-NEXT-7B uses $672{\times}672$ inputs and produces $N{=}2880$ visual tokens, five times the token count of LLaVA-1.5-7B.
This setting gives a stronger test of visual-token pruning because the language-model prefill cost grows with the visual prefix length.
\Cref{tab:main-llavanext} reports results at $K{=}320$, where each method keeps only $11.1\%$ of the visual tokens.
DiffPrune obtains $96.1\%$ average retention, the best result among the reported methods and $0.4\%$ higher than HoloV.
We use the same scorer and pruning recipe as in LLaVA-1.5-7B, with no resolution-specific scheduling or architectural change.

\paragraph{Results on Qwen2.5-VL-7B.}
Qwen2.5-VL-7B differs from the LLaVA family in vision encoder, projector and language backbone, and processes images at native resolution, so the visual-token count varies per image and results are reported by pruning rate rather than a fixed budget $K$.
Under the three pruning rates listed in \cref{tab:main-qwen}, DiffPrune leads every evaluated baseline across the five benchmarks, with the margin over HoloV widening from $4.0\%$ at $66.7\%$ to $6.1\%$ at $88.9\%$.
No architectural or hyperparameter adaptation is applied across backbones; the same Scorer, Soft Top-$K$, Throttler, Denoiser and schedule used on LLaVA-1.5-7B are reused without modification, supporting the claim that the framework is backbone-agnostic by construction.
We further examine this behavior across Qwen2.5-VL model scales in \cref{sec:appendix-scalability}.

\begin{table}[t]
  \centering
  \caption{
    \textbf{Results on Qwen2.5-VL-7B across three pruning rates.}
    Native-resolution inputs yield image-dependent token counts, so results are grouped by pruning rate; Avg.\ is normalized by the unpruned model.
  }
  \label{tab:main-qwen}
  \footnotesize
  \setlength{\tabcolsep}{3pt}
  \renewcommand{\arraystretch}{0.95}
  \begin{tabular}{lcccccc}
    \toprule
    Method & MMB & MME & POPE & SQA & VQA$^{\text{Text}}$ & Avg. \\
    \midrule

    \rowcolor{rowgray}
    \multicolumn{7}{c}{\textit{Upper Bound (100\%)}} \\
    Vanilla & 82.8 & 2304 & 86.1 & 84.7 & 84.8 & 100\% \\

    \midrule
    \rowcolor{rowgray}
    \multicolumn{7}{c}{\textit{Pruning Rate = 66.7\%}} \\
    FastV~{\scriptsize\textcolor{gray}{(ECCV'24)}} & 75.7 & 2072 & 82.2 & 78.5 & 77.9 & 92.3\% \\
    HoloV~{\scriptsize\textcolor{gray}{(NeurIPS'25)}} & 78.3 & 2093 & 85.0 & 79.8 & 78.9 & \textcolor{blue}{94.3}\% \\
    \rowcolor{rowblue}
    \textbf{DiffPrune (Ours)} & 81.7 & 2279 & 84.9 & 86.4 & 79.0 & \textcolor{red}{98.3}\% \\

    \midrule
    \rowcolor{rowgray}
    \multicolumn{7}{c}{\textit{Pruning Rate = 77.8\%}} \\
    FastV~{\scriptsize\textcolor{gray}{(ECCV'24)}} & 74.9 & 2036 & 80.7 & 78.0 & 69.0 & 89.2\% \\
    HoloV~{\scriptsize\textcolor{gray}{(NeurIPS'25)}} & 76.5 & 2043 & 82.3 & 79.8 & 70.3 & \textcolor{blue}{90.8}\% \\
    \rowcolor{rowblue}
    \textbf{DiffPrune (Ours)} & 81.0 & 2218 & 82.8 & 83.7 & 76.7 & \textcolor{red}{95.9}\% \\

    \midrule
    \rowcolor{rowgray}
    \multicolumn{7}{c}{\textit{Pruning Rate = 88.9\%}} \\
    FastV~{\scriptsize\textcolor{gray}{(ECCV'24)}} & 69.2 & 1940 & 78.6 & 77.4 & 60.3 & 84.3\% \\
    HoloV~{\scriptsize\textcolor{gray}{(NeurIPS'25)}} & 72.4 & 2006 & 80.7 & 79.5 & 61.8 & \textcolor{blue}{87.0}\% \\
    \rowcolor{rowblue}
    \textbf{DiffPrune (Ours)} & 76.7 & 2113 & 79.8 & 82.9 & 76.7 & \textcolor{red}{93.1}\% \\

    \bottomrule
  \end{tabular}
\end{table}

\subsection{Efficiency Analysis}
\label{sec:efficiency}

We profile all methods on LLaVA-1.5-7B with $336{\times}336$ inputs and $N{=}576$ visual tokens, using a single NVIDIA A6000 GPU, batch size one, and FP16 precision.
Latencies are averaged over $30$ forward passes after warm-up.
\Cref{tab:efficiency} reports TTFT at $K{=}64$, decomposed into vision encoding, pruning-module decision, and LLM prefill.


\begin{table}[t]
  \centering
  \caption{
    \textbf{Efficiency breakdown on LLaVA-1.5-7B} ($K{=}64$).
    Latencies are in ms; FLOPs are in T.
  }
  \label{tab:efficiency}
  \footnotesize
  \setlength{\tabcolsep}{3pt}
  \renewcommand{\arraystretch}{0.95}
  \begin{tabular}{lccccccc}
    \toprule
    Method & Loc. & $K$ & FLOPs (T) & Enc. & Pruner & Prefill & TTFT \\
    \midrule
    Full (LLaVA) & --  & 576 & 8.89 & 29.71 & 0.00 & 118.66 & 149.51 \\
    \midrule
    PruneSID & Pre & 64 & 2.13 & 31.80 & 43.39 & 39.74 & 115.84 \\
    PyramidDrop & L8--L24 & 64 & 4.89 & 29.54 & 5.02 & 70.08 & 105.90 \\
    HoloV & Pre & 64 & 2.15 & 31.96 & 2.77 & 41.48 & 77.39 \\
    \textbf{DiffPrune (Ours)} & Pre & 64 & 2.13 & 30.41 & \textbf{0.69} & 41.61 & \textbf{73.73} \\
    \bottomrule
  \end{tabular}
\end{table}

Vision encoding takes about $30$\,ms for all methods, since pruning is applied after the encoder.
The main differences come from the pruner and the LLM prefill.
PyramidDrop prunes inside the language model, so early layers still process the full visual prefix.
Its LLM prefill remains $70.08$\,ms with $4.89$\,T FLOPs.
In contrast, Pre-LLM methods reduce the visual prefix before it enters the language model, bringing LLM prefill to about $40$\,ms and FLOPs to $2.13$--$2.15$\,T.
Among these methods, the pruner itself becomes the main source of latency.
PRUNESID spends $43.39$\,ms on token selection, over $60\times$ the cost of DiffPrune, while HoloV spends $2.77$\,ms.
DiffPrune needs only $0.69$\,ms for selection and reaches the lowest TTFT in the comparison ($73.73$\,ms).
This follows directly from its inference path in \cref{eq:infer-fwd}: the Throttler and Denoiser are removed, leaving only one Scorer pass and index gathering.

\subsection{Ablation Studies}
\label{sec:ablations}

We isolate three training-time design choices in DiffPrune: the gradient-continuous Throttler, the VP-Noise form, and the diagonal Denoiser.
All variants keep the base VLM, Scorer architecture, training data, and schedule fixed.
\Cref{tab:ablation} reports average retention over GQA, MMBench, POPE, and MME on LLaVA-1.5-7B.

\begin{table}[t]
  \centering
  \caption{
    \textbf{Component ablation of DiffPrune on LLaVA-1.5-7B.}
    Variants replace one training-time component; scores are average retention over GQA, MMBench, POPE, and MME.
  }
  \label{tab:ablation}
  \footnotesize
  \setlength{\tabcolsep}{3pt}
  \renewcommand{\arraystretch}{0.95}
  \begin{tabular}{l l c c c}
    \toprule
    Variant & Replaced design & $K{=}64$ & $K{=}128$ & $K{=}192$ \\
    \midrule
    Full DiffPrune & -- & \textbf{95.5}\% & \textbf{97.3}\% & \textbf{98.3}\% \\
    \midrule
    Gumbel-STE Throttler & gradient-continuous & 87.4\% & 91.3\% & 94.0\% \\
    Scale-gate Throttler & VP-Noise form & 93.2\% & 94.8\% & 96.5\% \\
    Global-attention Denoiser & diagonal & 92.7\% & 95.0\% & 95.9\% \\
    \bottomrule
  \end{tabular}
\end{table}

Replacing the Throttler with a Gumbel-Softmax + STE variant gives the largest drop.
Retention decreases by $8.1$, $6.0$, and $4.3$ points at $K{=}64/128/192$, respectively.
Since the rest of the pipeline is unchanged, this result supports the claim from \cref{sec:pilot} that the gradient-continuous training path is important in the full VLM setting, not only in the DeiT probe.

The other two rows separate DiffPrune's internal design choices.
Replacing VP-Noise with the fully differentiable scale gate $\tilde{\mathbf{x}}_i=\alpha_i\mathbf{x}_i$ reduces retention by $2.3$, $2.5$, and $1.8$ points, showing that noise injection gives a stronger information restriction than simple rescaling.
Replacing diagonal attention in the Denoiser with global self-attention reduces retention by $2.8$, $2.3$, and $2.4$ points.
This supports per-token denoising, since global attention can let corrupted low-score tokens recover information from high-score tokens.

\subsection{Qualitative Analysis}
\label{sec:visualization}

\begin{figure}[h]
  \centering
  \includegraphics[width=\linewidth]{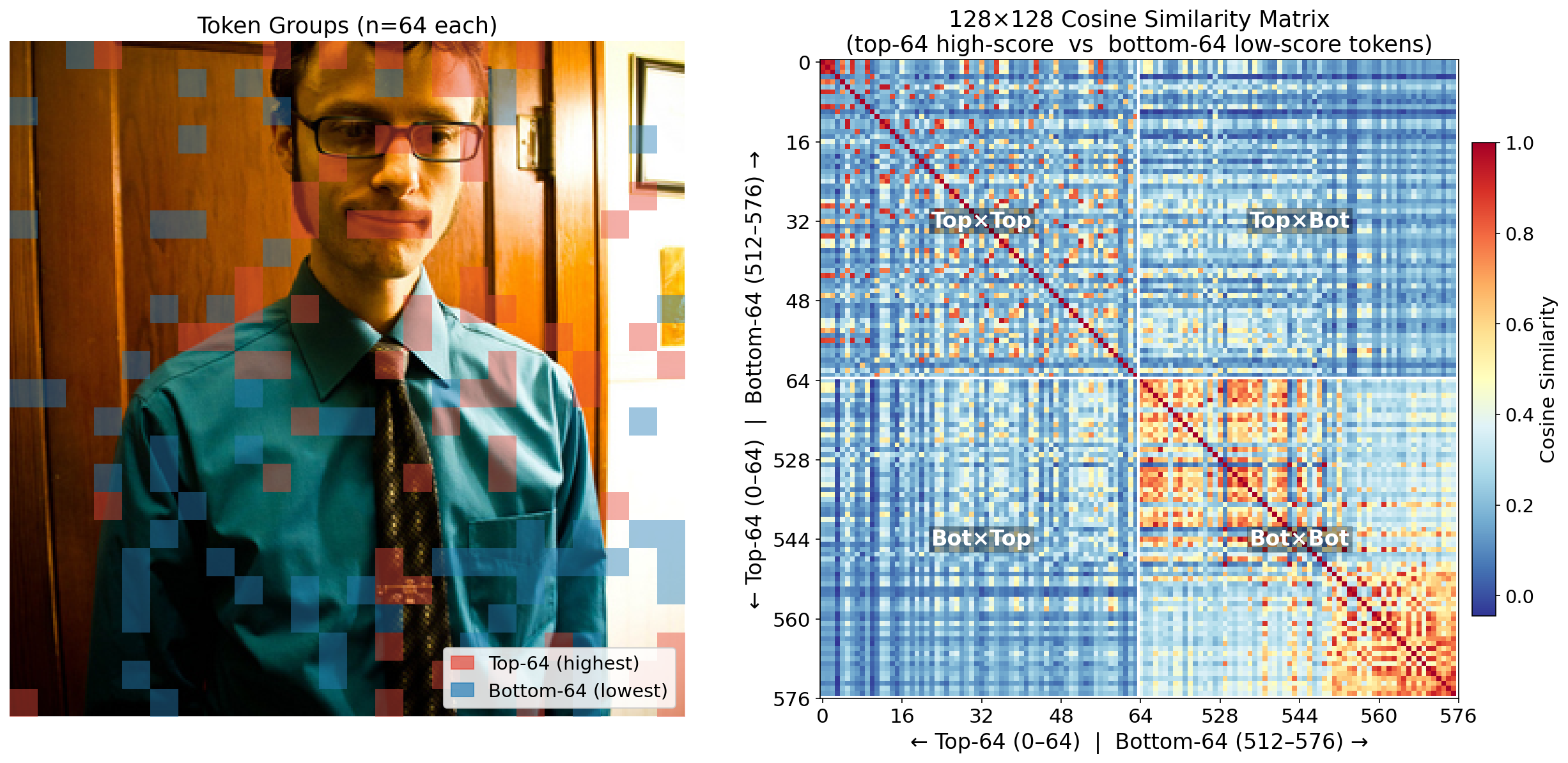}
  \caption{
\textbf{Learned token ranking on a VQAv2 example at $K{=}64$.} Left: Tokens are ranked by learned importance scores, where red and blue indicate the top-K retained tokens and the lowest-scored pruned tokens, respectively. Right: Similarity matrix of retained and pruned tokens. Pruned tokens are more redundant.}
  \label{fig:vqaviz}
\end{figure}

\Cref{fig:vqaviz} visualizes the token scores learned by DiffPrune on an example from the VQAv2 benchmark used in our evaluation.
For readability, we show the two extremes of the score ranking: the top-$64$ tokens cover visually salient regions such as the face, hands, and clothing, while the bottom-$64$ tokens mainly fall on visually uniform background areas.
The accompanying $128{\times}128$ cosine-similarity matrix $[\text{top-}64;|;\text{bottom-}64]$ provides a feature-space view of the same ranking: in this example, the retained-token block has lower internal similarity, whereas the lowest-scored block is more redundant.
This qualitative evidence is consistent with the quantitative results, while remaining illustrative rather than conclusive: the learned Scorer appears to allocate its limited budget toward visually informative and less redundant tokens.

\section{Conclusion}
\label{sec:conclusion}
We identify the discrete selection operator as a central obstacle in learned visual-token pruning: continuous importance scores must eventually become hard keep-or-drop decisions, and common surrogate-gradient solutions train through a backward path that is not the function executed in the forward pass.
DiffPrune addresses this mismatch by replacing training-time selection with continuous token-information control.
Its Soft Top-$K$ head and Information Throttler keep the scorer-to-loss path analytically differentiable during training, while inference collapses to a hard top-$K$ scorer-only path.
The resulting framework improves gradient-direction coherence by up to $28.4\times$, reduces local loss variation by roughly $24\times$ in our diagnostic study, and adds only $0.69$\,ms of selection overhead on LLaVA-1.5-7B.
These findings suggest that full differentiability is a useful design rule for learned pruning, not merely an implementation detail.
Extending the same principle to other sparsity operators remains future work.

\appendix

\setcounter{table}{0}
\renewcommand{\thetable}{A\arabic{table}}
\setcounter{figure}{0}
\renewcommand{\thefigure}{A\arabic{figure}}

\section{Implementation Details and Training Hyperparameters}
\label{sec:appendix-exp-setup}
\label{sec:appendix-impl}

\paragraph{Implementation.}
All experiments are implemented in PyTorch and run on a single NVIDIA RTX A6000 GPU.
The implementation is built on the LLaVA-v1.5 codebase~\cite{liu2023visual_instruction_tuning}.
For every evaluated backbone, the vision encoder, multimodal projector, and language model remain frozen; only the DiffPrune Scorer and Denoiser are optimized.

\paragraph{Architecture.}
The Scorer contains two Transformer encoder blocks with hidden dimension $1024$ and $16$ attention heads, followed by a linear projection that maps each visual token to a scalar importance logit.
The Denoiser is a single Transformer block with the diagonal attention mask described in \cref{sec:method-denoiser}.
Together, these two trainable modules add approximately $84$M parameters, about $1.2\%$ of LLaVA-1.5-7B.
At inference time, both the Throttler and Denoiser are removed; the deployed graph contains only the Scorer and hard top-$K$ gathering in \cref{eq:infer-fwd}.

\paragraph{Training.}
The Scorer and Denoiser are trained jointly in a single stage on a $10\%$ subset of the ImageNet-1K training split~\cite{deng2009imagenet}, paired with captions from ImageNet-1K-VL-Enriched~\cite{visual_layer_imagenet1k_vl_enriched}.
We use AdamW with learning rate $2{\times}10^{-4}$, cosine decay, and batch size $32$.
The objective is caption generation under the next-token negative log-likelihood in \cref{eq:train-fwd}; no labels from downstream evaluation benchmarks are used for training.
The Soft Top-$K$ temperature is annealed from $\tau_{\mathrm{start}}{=}2.0$ to $\tau_{\mathrm{end}}{=}0.1$ over $2{,}000$ steps, gradually sharpening the continuous retention weights toward the hard top-$K$ ranking used at inference.
Training runs for one epoch over the subset with early stopping based on validation loss and takes approximately ten GPU-hours.

\paragraph{Backbones and token budgets.}
LLaVA-1.5-7B uses $336{\times}336$ images and produces $N{=}576$ visual tokens.
We evaluate $K{\in}\{192,128,64\}$, corresponding to $66.7\%$, $77.8\%$, and $88.9\%$ pruning.
LLaVA-NEXT-7B uses $672{\times}672$ inputs and produces $N{=}2880$ visual tokens; we report $K{=}320$.
Qwen2.5-VL-7B processes images at native resolution, so the number of visual tokens varies by example; for this backbone, results are reported by pruning rate rather than by a fixed $K$.
Unless otherwise stated, the same Scorer, Soft Top-$K$ operator, Throttler, Denoiser, and training schedule are reused across backbones.

\paragraph{Baselines and reporting convention.}
Baselines include FastV~\cite{fastv}, SparseVLM~\cite{sparsevlm}, DART~\cite{dart}, PyramidDrop~\cite{pyramiddrop}, DivPrune~\cite{divprune}, VisionZip~\cite{visionzip}, HoloV~\cite{holov}, PRUNESID~\cite{fangprune}, ATP-LLaVA~\cite{atp-llava}, Dynamic-LLaVA~\cite{huang2025dynamic}, p-MoD~\cite{zhang2025pmod}, GlimpsePrune~\cite{zeng2025glimpse}, and GumbelDiffPrune. GumbelDiffPrune is an in-house controlled variant that replaces the DiffPrune Throttler with Gumbel-Softmax and a Straight-Through Estimator while keeping the rest of the pipeline fixed.
When applicable, Loc.\ records whether pruning happens before the language model (Pre) or inside it (L$a$--L$b$).
Entries are marked ``--'' when a compatible result is unavailable.

\paragraph{Efficiency profiling.}
Latency is profiled on LLaVA-1.5-7B with $336{\times}336$ inputs, $N{=}576$ visual tokens, batch size one, and FP16 precision on the same NVIDIA A6000 GPU.
We average $30$ forward passes after warm-up.
Time-to-first-token is decomposed into vision encoding, pruning-module decision time, and language-model prefill, so the reported pruning overhead isolates the cost added by each selector.

\section{Benchmark Descriptions}
\label{sec:appendix-benchmarks}

We evaluate DiffPrune on ten standard VLM benchmarks spanning compositional reasoning, perception, hallucination, text-centric understanding, and real-world VQA.

\paragraph{GQA~\cite{hudson2019gqa}.}
GQA is a compositional visual question answering benchmark built from Visual Genome scene graphs.
It emphasizes multi-step reasoning over objects, attributes, spatial relations, and logical operations, making it useful for testing whether token pruning preserves structured visual evidence.

\paragraph{MMBench and MMBench-CN~\cite{liu2024mmbench}.}
MMBench evaluates multimodal perception and reasoning through multiple-choice questions organized into fine-grained ability dimensions.
We report both the English benchmark and its Chinese counterpart, MMBench-CN, which follows the same evaluation taxonomy and enables testing under a language distribution shift.

\paragraph{MME~\cite{fu2023mme}.}
MME evaluates perception and cognition through manually designed yes/no questions.
Its perception subset covers skills such as existence, count, position, and color recognition, while its cognition subset covers broader reasoning categories.
Because each image is paired with controlled questions, MME is sensitive to both visual information loss and hallucinated content.

\paragraph{POPE~\cite{pope}.}
POPE measures object hallucination by querying whether specific objects are present in an image.
Questions are sampled under random, popular, and adversarial strategies, probing whether a model relies on visual evidence rather than dataset-level object co-occurrence priors.

\paragraph{ScienceQA-IMG~\cite{lu2022learn}.}
ScienceQA-IMG contains image-grounded science questions that require both visual understanding and domain knowledge.
We use the image-containing subset, following standard VLM evaluation practice.

\paragraph{VQAv2~\cite{goyal2017making}.}
VQAv2 is an open-ended visual question answering benchmark designed to reduce language priors through complementary image pairs.
It tests whether the model's answer changes with the visual evidence rather than being driven only by question statistics.

\paragraph{TextVQA~\cite{singh2019towards}.}
TextVQA focuses on reading and reasoning over text appearing inside images, such as signs, labels, documents, and scene text.
This benchmark stresses OCR-sensitive visual tokens and is therefore a stringent test for pruning methods that remove large fractions of the visual prefix.

\paragraph{SEED-Bench~\cite{li2023seed}.}
SEED-Bench evaluates image-centric multimodal comprehension across dimensions such as scene understanding, spatial relation, instance identity, and instance attribute recognition.
We use it to assess whether pruning preserves broad semantic coverage beyond conventional VQA accuracy.

\paragraph{VizWiz~\cite{bigham2010vizwiz}.}
VizWiz contains questions about images taken by blind users in everyday settings.
The images often contain unusual framing, low quality, or unanswerable queries, making the benchmark a realistic stress test for robustness under noisy visual inputs.

For each benchmark, we report its standard score.
To summarize performance across heterogeneous metrics, we additionally report average retention: each pruned score is normalized by the corresponding unpruned upper bound, and the normalized values are averaged across benchmarks.

\section{Scalability Across Model Scales}
\label{sec:appendix-scalability}

\begin{figure}[!htb]
  \centering
  \includegraphics[width=\textwidth]{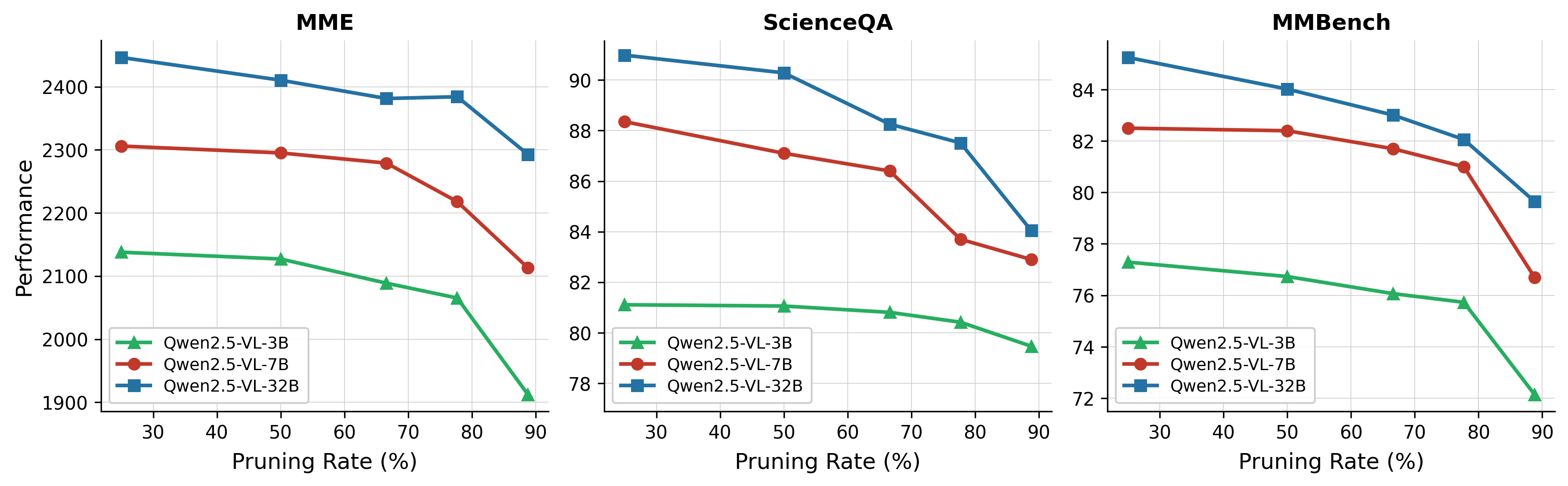}
  \caption{
    \textbf{Scalability across Qwen2.5-VL model scales.}
    Performance on MME, ScienceQA, and MMBench as a function of the visual-token pruning rate for Qwen2.5-VL-3B, 7B, and 32B.
  }
  \label{fig:appendix-scalability}
\end{figure}
To further test whether DiffPrune depends on a particular model capacity, we evaluate the 3B, 7B, and 32B variants of Qwen2.5-VL~\cite{bai2025qwen25vltechnicalreport} on MME, ScienceQA, and MMBench.
We use five visual-token pruning rates: $25\%$, $50\%$, $66.7\%$, $77.8\%$, and $88.9\%$.
The same pruning formulation and training recipe are used across model sizes.

\Cref{fig:appendix-scalability} shows that, across all three scales, performance decreases smoothly as the pruning rate increases, with no abrupt accuracy collapse even at $88.9\%$ pruning.
On MME, the 32B model remains above $2{,}400$ points near the $77.8\%$ pruning setting, while the 7B and 3B variants show steady rather than discontinuous degradation from their respective unpruned baselines.
MMBench and ScienceQA show the same pattern: the 32B and 7B results remain close through high pruning rates, and all models preserve usable performance under aggressive compression.
These results suggest that DiffPrune's budgeted, score-driven formulation is not tied to a single Qwen2.5-VL model size.

\section{Limitation and Future Work}

DiffPrune relies on the continuity of visual-token embeddings.
VP-Noise throttles a token by interpolating its feature vector with isotropic Gaussian noise under a variance-preserving schedule before the frozen language model consumes it.
This assumption is natural for VLM visual tokens, but it does not transfer directly to pure LLM settings, where the objects being pruned may be discrete text tokens, token-indexed KV entries, or weight structures whose semantics are not preserved under the same feature-space perturbation.
DiffPrune further introduces training-time cost through the Scorer and Denoiser, even though the Throttler and Denoiser are removed at inference.
Finally, while our pilot studies show substantially higher gradient-direction coherence and smoother scorer loss landscapes than Gumbel-Softmax pruning, a complete theoretical account of how feature dimensionality, the geometry of the $K$-th order statistic, and surrogate-gradient bias interact remains open.

Despite these limitations, these properties also suggest promising directions for future work in high-resolution visual domains. In particular, medical imaging and related settings are characterized by substantial redundancy in visual tokens~\cite{yang2025one,young2026fewer,young2026scalar}. This phenomenon has been widely observed in medical image analysis~\cite{yang2024segmentation,yang2023geometry,chen2026tc,wu2026multimodal}, multimodal medical foundation models~\cite{xu2023learning,xu2024medvilam,xu2024foundation,feng2026efficient}, and long-context visual reasoning tasks~\cite{gao2026zerosense,he2026autoselect}, suggesting that DiffPrune may be especially effective in these scenarios.

\bibliographystyle{unsrt}  
\bibliography{custom}  

\end{document}